\documentclass{article}
\usepackage{spconf,amsmath,graphicx}
\usepackage{tikz}
\usepackage{pgfplots}
\usepackage{graphicx}
\usepackage{bm}
\usepackage{epsfig}
\usepackage{amsmath}
\usepackage{amssymb}
\usepackage[ruled]{algorithm2e}
\usepackage{tabularx}
\usepackage{multirow}
\usepackage{booktabs}
\usepackage{pgfplots}
\pgfplotsset{width=7cm,compat=1.8}
\usepackage[
   singlelinecheck=off, labelfont =bf]{caption}
\usepackage{subfig}
\usepackage[colorlinks]{hyperref}
\usetikzlibrary{colorbrewer} 
\usepgfplotslibrary{colorbrewer}
\usepackage[utf8]{inputenc}
\usepackage[T1]{fontenc}
\usepackage{wrapfig}
\usepackage{lipsum}
\usepackage[misc]{ifsym}
\usepackage{cite}
\pgfplotsset{compat=1.11,
    /pgfplots/ybar legend/.style={
    /pgfplots/legend image code/.code={%
       \draw[##1,/tikz/.cd,yshift=-0.25em]
        (0cm,0cm) rectangle (3pt,0.8em);},
   },
}
\usepackage{enumitem}
\setlist{nosep, leftmargin=14pt}

\def\0{{\bf 0}}
\def\1{{\bf 1}}

\def\ie{{\em i.e.}}

\usepackage{mwe} 
\usepackage{booktabs}
\usepackage[normalem]{ulem}
\usepackage{lipsum} 
\useunder{\uline}{\ul}{}

\title{AEPL: Automated and Editable Prompt Learning for Brain Tumor Segmentation\thanks{This work has been submitted to the IEEE for possible publication. Copyright may be transferred without notice, after which this version may no longer be accessible.}}
\name{Yongheng~Sun$^{1,2}$, Mingxia~Liu$^{2,*}$, Chunfeng Lian$^{1,*}$}
\address{$^1$School of Mathematics and Statistics, Xi'an Jiaotong University, Xi'an 710049, China\\
$^2$Department of Radiology and BRIC, UNC at Chapel Hill, Chapel Hill, NC 27599, USA}
\begin{document}
%
\maketitle
\begin{abstract}
Brain tumor segmentation is crucial for accurate diagnosis and treatment planning, but the small size and irregular shape of tumors pose significant challenges. 
Existing methods often fail to effectively incorporate medical domain knowledge such as tumor grade, which correlates with tumor aggressiveness and morphology, providing critical insights for more accurate detection of tumor subregions during segmentation. 
We propose an Automated and Editable Prompt Learning (AEPL) framework that integrates tumor grade into the segmentation process by combining multi-task learning and prompt learning with automatic and editable prompt generation. 
Specifically, AEPL employs an encoder to extract image features for both tumor-grade prediction and segmentation mask generation. The predicted tumor grades serve as auto-generated prompts, guiding the decoder to produce precise segmentation masks. 
This eliminates the need for manual prompts while allowing clinicians to manually edit the auto-generated prompts to fine-tune the segmentation, enhancing both flexibility and precision. The proposed AEPL achieves state-of-the-art performance on the BraTS 2018 dataset, demonstrating its effectiveness and clinical potential. 
The source code can be accessed \href{https://github.com/YonghengSun1997/AEPL}{online}.

\end{abstract}

\begin{keywords}
Prompt Learning, Multi-Task Learning, Brain Tumor Segmentation
\end{keywords}
\section{Introduction}
\label{sec:intro}
Brain tumor segmentation plays a pivotal role in medical image analysis, serving as a cornerstone for accurate diagnosis and treatment planning. Precise delineation of tumor boundaries is critical for clinicians to assess tumor progression and design appropriate interventions. 
However, the inherent characteristics of brain tumors, such as their small sizes, irregular shapes, and often indistinct boundaries, 
pose significant challenges to automated tumor segmentation.
Traditional segmentation methods~\cite{norouzi2014medical} frequently struggle to handle these complexities, leading to suboptimal performance. 

Deep learning methods have recently shown great potential in automated medical image segmentation~\cite{ronneberger2015u, hatamizadeh2022unetr, xie2021cotr}. 
These approaches often fail to integrate medical domain expertise into the segmentation process. 
The absence of domain-specific knowledge can lead to less accurate tumor delineation, especially in challenging cases, where critical factors such as tumor grade—indicative of tumor aggressiveness and biological behavior—are essential for precise segmentation.  
Prompt learning~\cite{kirillov2023segment} has been used to guide the segmentation process by using specific input cues. 
However, existing techniques often require manual intervention, limiting their practical application in clinical environments where time and expertise may be constrained. 
To address these challenges, we propose Automated and Editable Prompt Learning (\textbf{AEPL}), a novel framework that introduces multi-task learning in conjunction with prompt learning. 
AEPL utilizes predicted tumor grades as automatically generated prompts, leveraging an encoder to extract image features for both segmentation and tumor grade prediction. 
These generated prompts are fed into a prompt encoder, guiding a U-Net decoder to produce segmentation masks, eliminating the need for manual prompt inputs. 
Importantly, our approach offers the flexibility of manual editing, allowing clinicians to refine the automatically generated prompts, thereby improving the precision of segmentation outcomes.

Our contributions are threefold. 
(1) An automated prompt learning strategy is introduced by seamlessly incorporating tumor-grade prediction into the segmentation process. This helps eliminate the need for manual prompts.  
(2) A novel editable prompt paradigm is designed,  enabling clinicians to interactively modify the generated prompts. 
(3) Extensive experiments performed on a benchmark dataset suggest the superiority of AEPL over several state-of-the-art (SOTA) methods, demonstrating its effectiveness and clinical utility.

\begin{figure*}[!th]
    \centering
 \setlength{\abovecaptionskip}{0pt}
\setlength{\belowcaptionskip}{0pt}
\setlength{\abovedisplayskip}{0pt}
\setlength{\belowdisplayskip}{0pt}   \includegraphics[width=0.98\linewidth]{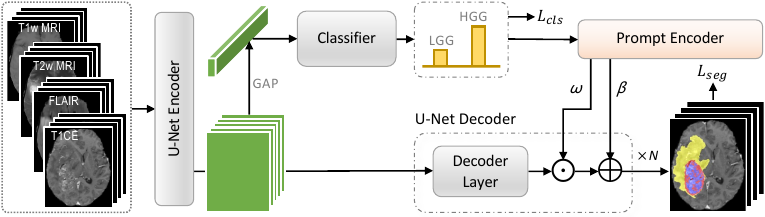} 
    \caption{The overall architecture of the proposed AEPL for brain tumor segmentation. Multi-modality MRI inputs are fed into the U-Net encoder to extract features. These features are then used to predict the tumor grade via classifier. The predicted tumor grade is subsequently fed as a prompt into a prompt encoder, guiding the U-Net decoder to produce precise segmentation masks. GAP: global average pooling; LGG: low-grade gliomas; HGG: high-grade gliomas.}
    \label{fig:pipeline}
\end{figure*}

\section{Methodology}
Our proposed AEPL is built upon a 3D U-Net backbone and consists of four key components: (1) a U-Net encoder, (2) tumor grade classifier, (3) a prompt encoder, and (4) a U-Net decoder, as illustrated in Fig.~\ref{fig:pipeline}. 
The AEPL is a fully end-to-end learning framework, with the training process divided into two main stages: feature extraction and multi-task learning stage, and prompt learning for segmentation stage. 
The two stages work together to ensure efficient feature extraction, tumor-grade prediction, and precise segmentation mask generation, all while allowing flexibility for manual intervention in the prompt learning phase.

\label{sec:format}

\subsection{Feature Extraction and Multi-Task Learning}
The AEPL processes multi-modality MRI inputs to simultaneously predict tumor grades and generate segmentation masks. 
Four modalities T1-weighted (T1w) MRI, T2w MRI, contrast-enhanced T1w (T1CE) MRI, and Flair MRI, are concatenated and fed into the U-Net encoder to extract high-dimensional representations. 
The extracted features are fed to the classifier to predict tumor grade and to the decoder to generate segmentation masks in a multi-task learning manner.

To integrate the segmentation task with tumor-grade prediction, the extracted features undergo global average pooling and are passed to a classifier, which predicts the tumor grade. 
The predicted grade and the ground truth are used to calculate a cross-entropy loss $L_{cls}$. 
The integration of classification and segmentation tasks enables the model to learn shared representations to boost the overall performance. 
This multi-task learning strategy ensures that the segmentation process benefits from tumor-grade information, enhancing the model's ability to differentiate between tumor types and shapes.

\subsection{Prompt Learning for Segmentation}
In AEPL, the tumor grades predicted by the classifier are used as \emph{automatically generated prompts} to guide the segmentation process.  
\if false
In AEPL, the predicted tumor grades serve a dual purpose. 
Besides being classification outputs, they also act as automatically generated prompts to guide the segmentation process. 
\fi 
Specifically, the prompt encoder inputs the tumor grade and outputs multi-scale weights $\omega=\{\omega_1, \dots, \omega_N\}$ and bias $\beta=\{\beta_1, \dots, \beta_N\}$, where $N$ is the number of upsampling in U-Net decoder. 
In each stage of the decoder, the output features are adjusted by multiplying the weight $\omega_i$ and adding the bias $\beta_i$. 
By incorporating tumor-grade information directly into the tumor segmentation task, AEPL leverages clinically relevant cues, allowing the model to produce more precise segmentation results.

Notably, the proposed AEPL also supports manual editing of automatically generated prompts, providing clinicians with the flexibility to fine-tune segmentation results by editing the predicted tumor grades. 
This editable prompt system provides a unique balance of automation and expert input, ensuring that the model's outputs can be adjusted based on clinical needs. 
This two-tier prompt learning mechanism not only reduces the need for manual input but also enhances the model’s flexibility, ensuring that it performs well across varying tumor shapes and sizes. 
The final loss function of the proposed AEPL consists of the segmentation loss and the classification loss, which is defined as:
\begin{equation}
    L=L_{seg}+\alpha*L_{cls}, 
    \label{eq_Loss}
\end{equation}
where $L_{seg}$ incorporates both the binary cross-entropy (BCE) loss and the Dice loss to optimize for both accuracy and overlap with ground-truth tumor masks, $L_{cls}$ represents the cross-entropy (CE) loss for tumor-grade classification (\ie, high-grade glioma vs. low-grade glioma), and $\alpha$ is a trade-off parameter. 
In our experiments, we empirically set the weighting factor $\alpha$ to 0.1 to balance the contributions of the segmentation and classification losses effectively. 
This hybrid loss ensures that the model learns to perform well on both tasks, enhancing overall performance in brain tumor segmentation.

\begin{table*}[]
\setlength{\abovecaptionskip}{0pt}
\setlength{\belowcaptionskip}{0pt}
\setlength{\abovedisplayskip}{0pt}
\setlength{\belowdisplayskip}{0pt}
\caption{Performance of different methods on BraTS 2018. The best results are in bold and second results are underlined.} 
\small
\centering
\renewcommand{\arraystretch}{0.9}
\label{tab:my-table}
\begin{tabular}{@{}l|llllll@{}}
\toprule
Method          & Dice\_ET               & Dice\_WT               & Dice\_TC               & HD95\_ET        & HD95\_WT        & HD95\_TC          \\ \midrule
U-Net     & 0.6472±0.2776          & 0.7375±0.2030          & 0.6649±0.3204          & 13.6479±48.3293        & 7.5647±9.8330          & 57.4925±127.7831         \\
TransFuse & 0.6292±0.2777          & 0.7264±0.1914          & 0.6959±0.2841          & 17.8532±49.0082        & 10.1073±14.5639        & 45.1285±113.2151         \\
TransUNet & 0.6532±0.2740          & 0.7432±0.1857          & 0.6681±0.3138          & 15.0208±48.7002        & 7.4019±11.4463         & 51.8809±120.8521         \\
CoTr      & 0.6646±0.2595          & 0.7449±0.2027          & 0.7418±0.2778          & 19.5668±67.5600        & 7.2068±9.2353          & 30.2129±94.5924          \\
nnFormer  & 0.6857±0.2473          & 0.7480±0.1924          & 0.7396±0.2790          & 12.8488±48.2909        & 12.6223±48.7384        & 30.3832±94.6137          \\
UNETR     & 0.6648±0.2507          & 0.7344±0.2063          & 0.6648±0.3214          & 19.3063±67.5407        & 7.6271±8.8478          & 50.0806±121.0686         \\
AEPL~(Ours)       & \underline{0.6892±0.2488} & \underline{0.7603±0.1936} & \underline{0.7752±0.2488} & \underline{7.3946±8.8380} & \underline{7.0219±11.4727}         & \underline{29.6881±94.6792} \\
AEPL-E~(Ours)       & \textbf{0.6898±0.2476} & \textbf{0.7604±0.1935} & \textbf{0.7758±0.2488} & \textbf{7.3700±8.8187} & \textbf{7.0210±11.4733}         & \textbf{29.6840±94.6810} \\ \bottomrule
\end{tabular}
\end{table*}

\section{Experiments}
\label{sec:illust}
\subsection{Experimental Setup}
The BraTS 2018 dataset, a widely used benchmark for brain tumor segmentation, is employed in this work.  
This dataset presents challenges due to variability in tumor size, shape, and location, making it ideal for developing robust segmentation models. 
It contains multi-modality MRI scans from 285 glioma patients. 
Each patient has four modalities: T1w, T2w, T1CE, and Flair MRI. 
Expert annotations are provided for three tumor regions: enhancing tumor (ET), tumor core (TC), and whole tumor (WT). 
The dataset includes high-grade gliomas (HGG) and low-grade gliomas (LGG), with ground truth available for the training set. 
In the proposed AEPL, the prompt is generated through the task of tumor-grade prediction (\ie, LGG vs. HGG classification).  

For a fair comparison, our method and the competing methods use the nnU-Net~\cite{isensee2021nnu} preprocessing pipeline, including cropping, resampling, normalization, and data augmentations. 
Our classifier and prompt encoder use a three-layer multi-layer perceptron (MLP), while the U-Net backbone consists of five downsampling and upsampling stages, each with two convolutional blocks. Deep supervision is applied with segmentation loss at five scales, weighted 0.0625, 0.125, 0.25, 0.5, and 1. SGD (momentum=0.99, weight decay=3e-5) is used for optimization, with an initial learning rate of 0.01 decayed to 1e-6 by polynomial learning rate decay schedule, batch size of 2, and patch size of (96, 160, 160). The dataset is split into training, validation, and test sets with a ratio of 6:2:2. The model is trained for 1,000 epochs, and the best-performing model on the validation set is selected for testing. We evaluate performance using volume Dice and 95th percentile Hausdorff distance (HD95). 
When one of the ground truth and prediction has no foreground, the conventional HD95 calculation fails. We set the HD95 value in this case to the diagonal length of 373 pixels.

\subsection{Results and Analysis}
\begin{figure*}[h]
\setlength{\abovecaptionskip}{0pt}
\setlength{\belowcaptionskip}{0pt}
\setlength{\abovedisplayskip}{0pt}
\setlength{\belowdisplayskip}{0pt}
    \centering
    \includegraphics[width=0.98\linewidth]{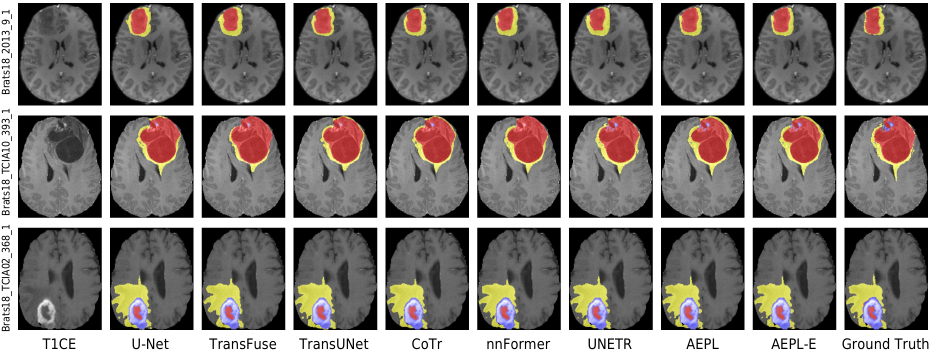} 
    \caption{Qualitative segmentation result. The whole tumor, enhancing tumor, and tumor core are shown in yellow, blue, and red, respectively. Our input contains four modalities, with only the contrast-enhanced T1w (T1CE) MRI input for illustration here.} 
    \label{fig:example}
\end{figure*}

We evaluate the performance of our AEPL  against several SOTA 2D and 3D segmentation methods. 
The 2D methods include \textbf{U-Net}~\cite{ronneberger2015u}, \textbf{TransFuse}~\cite{zhang2021transfuse}, and \textbf{TransUNet}~\cite{chen2021transunet}, while the 3D methods comprise \textbf{CoTr}~\cite{xie2021cotr}, \textbf{nnFormer}~\cite{zhou2021nnformer}, \textbf{UNETR}~\cite{hatamizadeh2022unetr}. 
To simulate clinical interaction, we replace the automatically generated prompts (through the classification task) with ground-truth tumor-grade labels, and denote this variant as \textbf{AEPL-E}. 
Table \ref{tab:my-table} presents a comparative analysis of segmentation performance across three tumor regions: enhancing tumor (ET), whole tumor (WT), and tumor core (TC). 
Our AEPL method consistently outperforms the competing methods in terms of Dice for all regions, achieving significant improvements, particularly in WT and TC regions. 
And AEPL achieves much lower HD95 values for ET and TC, indicating better boundary accuracy in challenging regions. 
Figure \ref{fig:example} further illustrates the qualitative results of tumor segmentation, from which we can see our methods provide more precise boundaries compared to other approaches. 
The visualized outputs demonstrate that AEPL effectively addresses the segmentation of small and irregularly shaped tumors, producing more accurate masks.

To investigate the influence of clinical intervention via prompt, we further study the intermediate results (used as prompt) of tumor-grade prediction in AEPL. 
We find that our AEPL achieves an accuracy of 96.75\% in LGG vs. HGG classification. 
As shown in Table \ref{tab:my-table} and Fig. \ref{fig:example}, AEPL (with generated tumor grades as prompts) is slightly inferior to AEPL-E which uses the manually annotated ground-truth labels as prompts. 
This implies that using editable prompts via clinical intervention (as we do in AEPL-E) can boost segmentation performance, providing flexibility for manual refinement for enhanced clinical applicability. 
Additionally, the results of AEPL-E and AEPL are not significantly different (with $p > 0.05$ via paired $t$-test). 
This is possibly due to the high accuracy (\ie, 96.75\%) of AEPL in tumor-grade prediction. 

\if false 
As shown in Table \ref{tab:my-table} and Fig. \ref{fig:example}, this intervention led to improvements across all segmentation metrics. 
The integration of automatic and editable prompt generation not only improves segmentation performance but also provides flexibility for manual refinement, further enhancing clinical applicability.
However, given the high classification accuracy, the magnitude of improvement from clinical intervention was relatively modest.
\fi

\textbf{\begin{table}[]
\setlength{\abovecaptionskip}{0pt}
\setlength{\belowcaptionskip}{0pt}
\setlength{\abovedisplayskip}{0pt}
\setlength{\belowdisplayskip}{0pt}
\caption{Results of AEPL with different  $\alpha$ on BraTS 2018. The best results are in bold and second results are underlined.}
\renewcommand{\arraystretch}{0.9}
\label{tab:alpha}
\resizebox{\columnwidth}{!}{%
\begin{tabular}{@{}l|llllll@{}}
\toprule
   $\alpha$  & Dice\_ET        & Dice\_WT        & Dice\_TC        & HD95\_ET        & HD95\_WT        & HD95\_TC         \\ \midrule
0    & 0.6788          & 0.7596          & {\ul 0.7689}    & 14.0289         &{\ul 6.3836}          & \textbf{28.9325} \\
0.01 & 0.6836          & 0.7588          & 0.7571          & 13.0344         & 7.1064          & 35.3232          \\
0.1  & {\ul 0.6892}    & 0.7603          & \textbf{0.7752} & \textbf{7.3946} & 7.0219    & {\ul 29.6881}    \\
1    & 0.6819          & {\ul 0.7629}    & 0.7058          & 18.8810         & \textbf{6.1324} & 55.9449          \\
10   & \textbf{0.6907} & \textbf{0.7637} & 0.7020          & {\ul 12.5831}   & 12.0047         & 55.3643          \\ \bottomrule
\end{tabular}%
}
\end{table}
}


\subsection{Parameter Analysis} 
We further study the influence of $\alpha$ in Eq.~\eqref{eq_Loss} by varying its values within the range of \{0, 0.01, 0.1, 1, 10\} and report the results of AEPL in Table~\ref{tab:alpha}. 
The results demonstrate that our method with $\alpha=0.1$ yields the overall best performance, implying that segmentation loss contributes more to the final results. 
But without the classification task (\ie, $\alpha=0$), AEPL can not generate good performance, further validating the necessity of classification-based prompt input. 

\if false
indicating an optimal balance between the multi-task learning objective. 
Lower values of $\alpha$ lead to under-emphasis on the classification task, while higher values cause a degradation in segmentation accuracy due to over-penalization. 
\fi 

\section{Conclusion}
This paper presents a brain tumor segmentation framework, called AEPL, with multi-task learning and prompt learning. AEPL utilizes predicted tumor grades as automatically generated prompts, guiding the segmentation process without the need for manual intervention. It also offers the flexibility to edit prompts for enhanced segmentation performance. 
Extensive evaluations on a public dataset suggests the superiority of AEPL over state-of-the-art methods. 
Our method not only provides a robust, adaptable tool for accurate brain tumor segmentation with enhanced clinical applicability.

\section{Compliance with Ethical Standards}
\label{sec:ethics}
This research study was conducted retrospectively using human subject data made available in open access by the BraTS 2018 Challenge. Ethical approval was not required as confirmed by the license attached with the open access data.
      
      \section{Acknowledgments}
\label{sec:acknowledgments}
The research of C.~Lian was supported in by NSFC Grants (Nos.~12326616, 62101431, and 62101430) and Natural Science Basic Research Program of Shaanxi (No. 2024JC-TBZC-09).  
The research of M.~Liu was supported by NIH Grant (No.~AG073297). 



\bibliographystyle{IEEEbib}
\bibliography{AEPL}

\end{document}